\documentclass[conference]{IEEEtran}
\IEEEoverridecommandlockouts

\newif\iffinal
\finaltrue

\newif\ifieee
\ieeefalse

\usepackage{cite}
\usepackage{amsmath,amssymb,amsfonts}
\usepackage{algorithmic}
\usepackage{graphicx}
\usepackage{textcomp}

\usepackage[caption=false,farskip=0pt]{subfig}
\usepackage{booktabs}
\usepackage{multirow}
\usepackage[hyphens]{url}
\usepackage{balance}

\ifieee
\else
\usepackage[colorlinks=true,allcolors=black]{hyperref} %
\fi

\usepackage[acronym]{glossaries}
\usepackage{xspace}
\usepackage[utf8]{inputenc}
\usepackage[T1]{fontenc}
\usepackage[dvipsnames,table]{xcolor}
\usepackage{diagbox}
\usepackage[capitalise]{cleveref} %
\usepackage{comment}

\usepackage{soul}

\newcounter{fncounter}
\newcommand*{\RL}[2][]{\textcolor{Rhodamine}{[\textbf{\ifthenelse{\equal{#1}{}}{RL}{RL(#1)}}: #2]}}
\newcommand\RLI[1]{} %
\newcommand*{\DM}[2][]{\textcolor{blue}{[\textbf{\ifthenelse{\equal{#1}{}}{DM}{DM(#1)}}: #2]}}
\newcommand*{\VE}[2][]{\textcolor{ForestGreen}{[\textbf{\ifthenelse{\equal{#1}{}}{VE}{VE(#1)}}: #2]}}

\newacronymstyle{long-short-br}
{%
  \GlsUseAcrEntryDispStyle{long-short}%
}%
{%
  \GlsUseAcrStyleDefs{long-short}%
}
\setacronymstyle{long-short-br}

\newcommand\major[1]{{#1}} %

\makeatletter
\def\footnoterule{\kern-3\p@
  \hrule \@width 2in \kern 2.6\p@} %
\makeatother

\let\OLDthebibliography\thebibliography
\renewcommand\thebibliography[1]{
  \OLDthebibliography{#1}
  \setlength{\itemsep}{2pt}
}

\begin{document}

\newacronym{cnn}{CNN}{Convolutional Neural Network}
\newacronym{hr}{HR}{high-resolution}
\newacronym{lp}{LP}{license plate}
\newacronym{lpr}{LPR}{License Plate Recognition}
\newacronym{lr}{LR}{low-resolution}
\newacronym{misr}{MISR}{Multi-Image Super-Resolution}
\newacronym{mse}{MSE}{Mean Squared Error}
\newacronym{ocr}{OCR}{Optical Character Recognition}
\newacronym{psnr}{PSNR}{Peak Signal-to-Noise Ratio}
\newacronym{sisr}{SISR}{Single-Image Super-Resolution}
\newacronym{srcnn}{SRCNN}{Super-Resolution Convolutional Neural Network}
\newacronym{ssim}{SSIM}{Structural Similarity Index Measure}
\newacronym{vsr}{VSR}{Video Super-Resolution}
\newacronym{rcb}{RCB}{Residual Concatenation Block}
\newacronym{fm}{FM}{Feature Module}
\newacronym{sfe}{SFE}{Shallow Feature Extractor}
\newacronym{rm}{RM}{Reconstruction Module}
\newacronym{ps}{PS}{\textit{PixelShuffle}}
\newacronym{pu}{PU}{\textit{PixelUnshuffle}}
\newacronym{nn}{NN}{Neural Network}
\newacronym{psfe}{PSFE}{Pre-shallow Feature Extractor}
\newacronym{ca}{CA}{Channel Unit}
\newacronym{pos}{POS}{Positional Unit}
\newacronym{tfam}{TFAM}{Two-fold Attention Module}
\newacronym{map}{MAP}{Maximum a Posteriori}
\newacronym{gan}{GAN}{Generative Adversarial Network}
\newacronym{ccpd}{CCPD}{Chinese City Parking Dataset}
\newacronym{mprnet}{MPRNet}{Multi-Path Residual Network}
\newacronym{cbam}{CBAM}{Convolution Block Attention Module}
\newacronym{se}{SE}{Squeeze-and-excitation}
\newacronym{esa}{ESA}{Enhanced Spatial Attention}
\newacronym{csrgan}{CSRGAN}{Character-Based Super-Resolution Generative Adversarial Networks}
\newacronym{dconv}{DConv}{depthwise-separable convolutional layer}
\newacronym{gp}{GP}{Geometrical Perception Unit}
\newacronym{srgan}{SRGAN}{Super-Resolution Generative Adversarial Networks}
\newacronym{dpca}{DPCA}{Dual-Coordinate Direction Perception Attention}
\newacronym{dganesr}{D\textunderscore GAN\textunderscore ESR}{Double Generative Adversarial Networks for Image Enhancement and Super Resolution}
\newacronym{rdb}{RDB}{Residual Dense Block}
\newacronym{pltfam}{PLTFAM}{Pixel Level Three-Fold Attention Module}
\newacronym{bnn}{BNN}{Bayesian Neural Network}
\newacronym{sibgrapi}{SIBGRAPI}{Conference on Graphics, Patterns and Images}

\newcommand{\pku}{PKU\xspace}
\newcommand{\rodosol}{RodoSol-ALPR\xspace}
\newcommand{\ufpr}{UFPR-ALPR\xspace}
\newcommand{\supplementary}{\url{https://github.com/valfride/lpr-rsr-ext/}}

\newcommand{\indentresults}{\hspace{3mm}}

\iffinal
\title{\huge Super-Resolution of License Plate Images Using\\Attention Modules and Sub-Pixel Convolution Layers
}
\else
\title{Super-Resolution of License Plate Images Using Attention Modules and Sub-Pixel Convolution Layers
\thanks{The funding agencies are hidden for review.}
}
\fi

\iffinal
\author{Valfride Nascimento\IEEEauthorrefmark{1}, Rayson Laroca\IEEEauthorrefmark{1}, Jorge de A. Lambert\IEEEauthorrefmark{2}, William Robson Schwartz\IEEEauthorrefmark{3}, and David Menotti\IEEEauthorrefmark{1}\\
\IEEEauthorrefmark{1}\hspace{0.2mm}\normalsize Department of Informatics, Federal University of Paran\'a, Curitiba, Brazil\\
\IEEEauthorrefmark{2}\hspace{0.2mm}\normalsize Regional Superintendence at Bahia, Brazilian Federal Police, Salvador, Brazil\\
\IEEEauthorrefmark{3}\hspace{0.2mm}\normalsize Department of Computer Science, Federal University of Minas Gerais, Belo Horizonte, Brazil\\
\resizebox{0.9\linewidth}{!}{
\IEEEauthorrefmark{1}{\tt\small \{vwnascimento,rblsantos,menotti\}@inf.ufpr.br} \quad \IEEEauthorrefmark{2}{\tt\small lambert.jal@pf.gov.br} \quad \IEEEauthorrefmark{3}\hspace{0.3mm}{\tt\small william@dcc.ufmg.br}
}\thanks{This manuscript is a postprint of a paper accepted by \textit{Computers \& Graphics}. See the final version on \textit{Science Direct} (DOI: \href{https://doi.org/10.1016/j.cag.2023.05.005}{\textcolor{blue}{10.1016/j.cag.2023.05.005}}).}}

\maketitle

\begin{abstract}
Recent years have seen significant developments in the field of \gls*{lpr} through the integration of deep learning techniques and the increasing availability of training data.
Nevertheless, reconstructing \glspl*{lp} from \gls*{lr} surveillance footage remains challenging.
To address this issue, we introduce a \gls*{sisr} approach that integrates attention and transformer modules to enhance the detection of structural and textural features in \gls*{lr} images.
Our approach incorporates \textit{sub-pixel convolution layers} (also known as PixelShuffle) and a loss function that uses an \gls*{ocr} model for feature extraction.
We trained the proposed architecture on synthetic images created by applying heavy Gaussian noise to high-resolution \gls*{lp} images from two public datasets, followed by bicubic downsampling.
As a result, the generated images have a \gls*{ssim} of less than 0.10.
Our results show that our approach for reconstructing these low-resolution synthesized images outperforms existing ones in both quantitative and qualitative measures.
Our code is publicly available at \supplementary.
\end{abstract}

\glsresetall
\section{Introduction}

Super-resolution is a method for enhancing the quality of an image or video by increasing its resolution.
It has become a widespread technology in fields like medical imaging and surveillance~\cite{yue2016image, liu2023blind}.
In recent times, there have been remarkable advancements in super-resolution techniques, particularly in interpolation-based, example-based, and deep learning-based methods~\cite{wang2021deep, zhang2021residual,santos2022face}.
These improvements have made it feasible to enhance \gls*{lr} images and videos in a manner that was once considered~impossible.

Despite advances in recent years, super-resolution remains a challenging issue due to its ill-posed nature, where there can be numerous solutions in the \gls*{hr} space~\cite{wang2021deep, liu2023blind}.
Furthermore, the computational difficulty of the problem grows as the upscale factor increases, and \gls*{lr} images may lack sufficient information to reconstruct the desired details~\cite{wang2021deep, liu2023blind}.
Super-resolution can be classified into three main categories: \gls*{sisr}, \gls*{misr}, and video super-resolution~\cite{guarnieri2021perspective, liu2023blind}.
This study focuses on the application of \gls*{sisr} in the context of \gls*{lpr}, as images from real-world surveillance systems are often characterized by low resolution and poor quality~\cite{goncalves2019multitask,maier2022reliability,moussa2022forensic}.
Although such challenging conditions are common in forensic applications, recent studies in \gls*{lpr} have mainly concentrated on scenarios where the \glspl*{lp} are perfectly legible~\cite{laroca2021efficient,gong2022unified,silva2022flexible,wang2022rethinking}.

To address the super-resolution problem, \major{many researchers have proposed approaches based on \glspl*{cnn}~\cite{lucas2019generative,mehri2021mprnet,liu2023blind}}.
These approaches have achieved exceptional results, but often rely on deep architectures that can be computationally expensive and focus on increasing the \gls*{psnr} and \gls*{ssim} without considering the particular application at hand.
In the context of \gls*{lpr}, we assert that such methods may not be effective in dealing with confusion between characters, such as `Q' and `O', `T' and `7', and `Z' and `2', which can pose a problem in~\gls*{lpr}.

We present a novel approach for improving \gls*{lp} super-resolution through the use of \gls*{ps} layers and a Three-Fold Attention Module.
Our method extends the work of Mehri et al.~\cite{mehri2021mprnet} and Nascimento et al.~\cite{nascimento2022combining} by taking into account not only the pixel intensity values, but also structural and textural information. 
To further enhance the performance, we incorporate an auto-encoder that extracts shallow features by squeezing and expanding the network constructed with \gls{ps} and \gls{pu} layers. 
Additionally, we leverage a pre-trained \gls*{ocr} model~\cite{goncalves2018realtime} to extract features from the \gls*{lp} images during the training phase, resulting in improved super-resolution performance and recognition rates. 
It is notable that the choice of the \gls*{ocr} model can be tailored to specific application~requirements.

In summary, the main contributions of this work are:
\begin{itemize}
\item A super-resolution approach that builds upon \acrshort{mprnet}~\cite{mehri2021mprnet} and the architecture we proposed in~\cite{nascimento2022combining} (see the next paragraph) by incorporating subpixel-convolution layers (\gls*{ps} and \gls*{pu}) in combination with a \gls{pltfam};
\item A novel perceptual loss that combines features extracted by an \gls*{ocr} model~\cite{goncalves2018realtime} with L1 loss to reconstruct characters with the most relevant characteristics.
This loss function allows the use of any \gls*{ocr} model for~\gls*{lpr};
\item The datasets we built for this work, as well as the source code, are publicly available to the research community.
\end{itemize}

A preliminary version of this study was presented at the 2022 \gls*{sibgrapi}~\cite{nascimento2022combining}.
The approach described here differs from the previous version in several aspects.
For example, we introduce novel approaches for \gls*{lp} super-resolution, such as an attention module architecture that considers vertical and horizontal lines to extract more structural and textural details of the \gls*{lp} font.
We propose a new loss method that employs feature extraction through a pre-trained network for \gls*{lp} recognition.
The images used for training and testing consist of paired low- and high-resolution \glspl*{lp}, with the \gls*{lr} samples degraded until their \gls*{ssim} falls below 0.10. 
These improvements have enabled us to achieve better results than those reported in our previous work.
In this work, we report the results of experiments performed on two datasets, collected in different regions under various conditions, instead of a single one.
In the \rodosol dataset~\cite{laroca2022cross}, our approach recognizes at least five characters in $74.2\%$ of \glspl*{lp} compared to $42.2\%$ by our previous model trained and evaluated under the same conditions.
In the \pku dataset~\cite{yuan2017robust}, the improvement was even more significant, from $82.5$\% by the preliminary approach to $97.3$\% by the improved one (proposed in this~work).

The rest of this article is structured as follows. \cref{RelatedWork} provides a concise overview of relevant studies on \gls*{sisr}, as well as works that designed or applied super-resolution techniques specifically to \gls*{lpr}.
In~\cref{Methodology}, we elaborate on our proposed network architecture and the implementation of the new perceptual loss function that explores an \gls*{ocr} model as a feature extractor.
\cref{Experiments} presents the experiments performed and the results obtained.
In \cref{Conclusion}, we summarize the findings and their significance, concluding this~study.
\section{Related Work}
\label{RelatedWork}

This section provides a brief overview of related work.
Some approaches used in \gls*{sisr} are discussed in greater detail in \cref{sisr}, and the use of deep learning methods for \gls*{lp} super-resolution is discussed in Section 2.2.

\subsection{Single-Image Super-Resolution}

\label{sisr}

The field of \gls*{sisr} has experienced significant advancements in recent years, leading to its broad application in various domains~\cite{yue2016image, liu2023blind}.
Early \gls*{sisr} methods were generally classified into four categories: prediction models, edge-based methods, image statistical methods, and example-based methods~\cite{glasner2009super, kim2010single, timofte2013anchored, yang2013fast, yang2014single}.
In 2016, Dong et al.~\cite{dong2016image} introduced the \gls*{srcnn}, a deep learning-based approach to \gls*{sisr}, which demonstrated both superior quality and faster performance compared to previous~methods.

\major{Despite the success of \gls*{srcnn}, some limitations were observed such as} relying on pre-upsampled \gls*{lr} images, which drastically increased computational complexity without providing significant additional information for image restoration~\cite{wang2015deep, chen2017trainable}.
To overcome these limitations, later studies by Dong et al.~\cite{dong2016accelerating} and Shi et al.~\cite{shi2016realtime} incorporated the upsampling process near the end of the network architecture, \major{leading to a substantial reduction in execution time, parameters, and computational cost.}

Shi et al.~\cite{shi2016realtime} highlighted the importance of learnable upscaling and designed specialized sub-pixel convolution layers to optimize the learning of upscaling filters.
This enabled the networks to learn complex mappings from \gls*{lr} to \gls*{hr} images, resulting in improved performance compared to using fixed filters from interpolation~methods.

Recent research in the field of super-resolution has introduced attention mechanisms as a means of improving image reconstruction.
Zhang et al.~\cite{zhang2018image} were among the pioneers to introduce the use of first-order statistical attention mechanisms in this context.
Afterwards, Dai et al.~\cite{dai2019second} presented an improved version that uses second-order statistics to extract more meaningful features.
Huang et al.~\cite{huang2021interpretable} proposed an attention network that preserves detail fidelity by using a divide-and-conquer strategy to progressively process smooth and detailed~features.

Mehri et al.~\cite{mehri2021mprnet} introduced the \gls*{mprnet}, which leverages information from both inner-channel and spatial features using a \gls*{tfam}.
\gls*{mprnet} has demonstrated superior or competitive performance compared to multiple state-of-the-art methods such as those presented in~\cite{he2019ode, luo2020latticenet, muqeet2020ultra}.

Recently, Zhang et al.~\cite{zhang2023single} proposed a structure- and texture-preserving image super-resolution reconstruction method, known as the \gls*{dpca} mechanism.
This method effectively emphasizes structure and feature details, resulting in improved image quality compared to previous~methods.

\subsection{Super-Resolution for License Plate Recognition}
\label{srOnLp}

The goal of \gls*{lpr} is to precisely extract and identify characters from each \gls*{lp}.
Despite recent progress and successful outcomes in LPR~\cite{laroca2021efficient, wang2022rethinking, silva2022flexible}, most of the models proposed have only been trained and evaluated on \gls*{hr} images, where the \gls*{lp} characters are clear and easily recognizable to the human eye.
This does not reflect the typical conditions encountered in real-world surveillance scenarios, where images frequently have low resolution and poor~quality~\cite{goncalves2019multitask,maier2022reliability,moussa2022forensic}.

The quality of \gls*{lp} images is closely linked to various factors, such as camera distance, motion blur, lighting conditions, and image compression techniques used for storage~\cite{goncalves2019multitask}.
In commercial \gls*{lpr} systems, sharp images are typically captured using global shutter cameras. However, in surveillance systems, cost-effective cameras that use rolling shutter technology are often employed, leading to blurry images~\cite{liang2008analysis} with illegible~\glspl*{lp}.

Super-resolution techniques have been proposed as a solution to address the issue of poor image quality in \gls*{lpr}.
The first works to combine the idea of super-resolution and \gls*{lp} recognition date back to the 2000s, such as those proposed by Suresh et al.~\cite{suresh2007superresolution} and Yuan et al.~\cite{yuan2008fast}, which rely on image processing or interpolation techniques to increase resolution.
While some algorithms incorporate character semantics and segmentation for super-resolution, these methods tend to perform poorly on noisy images~\cite{zou2019semantic}. Considering space limitations, the remainder of this section describes works published in recent years.

Lin et al.~\cite{lin2021license} proposed a super-resolution approach for \gls*{lpr} using their \gls*{srgan} model and a perceptual \gls*{ocr} loss. 
Despite obtaining promising results, the experiments were limited to $100$ images and excluded those with low brightness/contrast.

Hamdi et al.~\cite{hamdi2021new} proposed a GAN-based architecture named \gls*{dganesr}, which outperformed previous \gls*{srgan} methods~\cite{lin2021license}.
The architecture consists of two networks concatenated together, with the first network responsible for denoising and deblurring and the second network performing super-resolution. 
The authors assessed the performance of their method in terms of \gls*{psnr} and \gls*{ssim}, but acknowledged that these metrics alone do not necessarily indicate superior image reconstruction.
The model was trained using \gls*{lr} images downsampled from \gls*{hr}~images.

Lee et al.~\cite{lee2020super} observed that previous super-resolution approaches did not take character recognition into account.
Thus, they designed a GAN-based model that incorporates a perceptual loss composed of intermediate features extracted by a scene text recognition model.
Specifically, the authors used an intermediate representation (\textit{block4}) of ASTER~\cite{shi2019aster}.
While their method produced better results than the same GAN-based model trained with the original perceptual loss, the authors did not make the dataset used available, and the degradation method employed was not~detailed.

Although the primary objective of enhancing \gls*{lp} images is to improve recognition accuracy, it is surprising that most works have primarily evaluated the quality of the reconstructed images through subjective visual evaluations or metrics such as \gls*{psnr} and \gls*{ssim}.
It is well-known that these metrics have a limited correlation with human assessment of visual quality~\cite{johnson2016perceptual,zhang2018unreasonable}.
Furthermore, we observed that most previous studies explored private datasets in the experiments~\cite{svoboda2016cnn,lee2020super,hamdi2021new,maier2022reliability}, which makes it challenging to accurately assess the reported~results.
\section{Proposed Approach}
\label{Methodology}

\begin{figure*}[!htb]
\centering
    \includegraphics[width=0.99\linewidth]{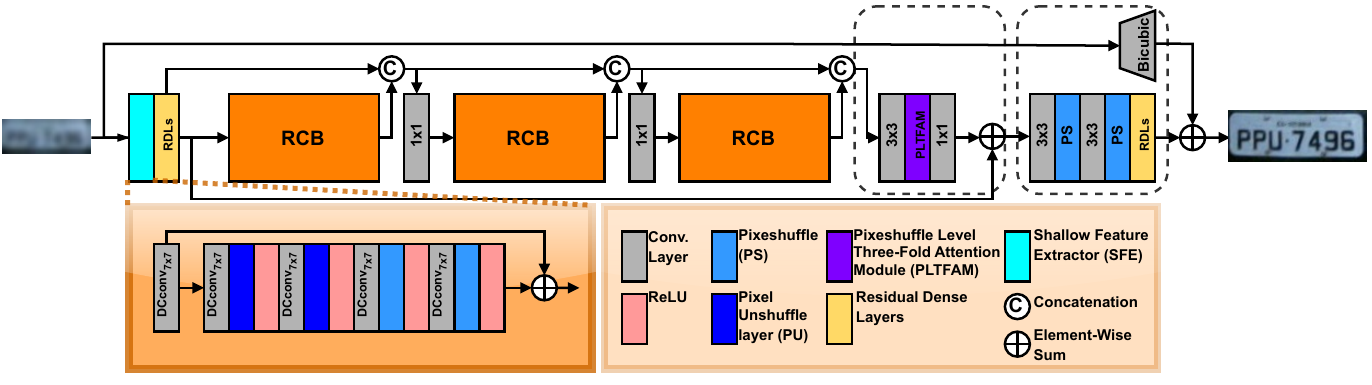}
    
    \vspace{-3.75mm}
    
    \caption{
    The proposed architecture, which incorporates an autoencoder consisting of \gls*{ps} and \gls*{pu} layers for feature compression and expansion, respectively. The aim of this design is to eliminate less significant features.
    In addition, the \gls*{tfam} modules in the original architecture were replaced with \gls*{pltfam} modules throughout the network. The legend inside the figure provides explanations for the acronyms~used.
    }
    \label{fig:nnarch}
\end{figure*}

This section details our super-resolution approach that enhances the extraction of structural and textural features from low-resolution \glspl{lp}.
Our network extends the network proposed in our previous work~\cite{nascimento2022combining}, further expanding the \gls{mprnet} architecture and \gls{tfam} algorithm by Mehri et al.~\cite{mehri2021mprnet} while taking inspiration from~\cite{zhang2023single} to improve the proposed attention module to enable the network to capture structural and textural information.
The proposed approach leverages a novel perceptual loss function that uses an \gls{ocr} model as a feature~extractor.

\subsection{Network Architecture Modifications}

The proposed approach for super-resolution in \gls*{lpr} features a network architecture that builds upon the work of Mehri et al.~\cite{mehri2021mprnet} and Zhang et al.~\cite{zhang2023single}.
As illustrated in \cref{fig:nnarch}, the architecture comprises four key components: a \gls*{sfe}; \glspl*{rdb} (refer to \cite{zhang2021residual} for more information); a \gls*{fm} module; and a \gls*{rm}. The \gls*{rm} combines the output of the \gls*{fm} module with two long-skip connections, one from the end of the \gls*{sfe} module and the other from the input image, to produce the final high-resolution output.
Our specific modifications are discussed in the following~paragraphs.

The design of the \gls*{sfe} block includes a convolutional layer with a $5\times5$ kernel, followed by an autoencoder that employs \glspl*{dconv}, \gls*{pu} and \gls*{ps} operations instead of conventional convolutional layers and pooling and upscale operations.
The output of the layers is then combined with a skip connection from the initial convolutional layers.
Finally, the resulting output is processed by the~\glspl*{rdb}.

In \cref{fig:pltfam}, we show our modifications to the \gls{mprnet}'s \gls{tfam}~\cite{mehri2021mprnet} and to the attention module by Nascimento et al.~\cite{nascimento2022combining} to create the \gls*{pltfam}.
The design of this module is based on the following insights: \textbf{(i)}~images are composed of the relationship between channels, where each channel contributes unique characteristics to form the final image, therefore, the extraction of these features is crucial for proper image restoration; \textbf{(ii)}~the positional information of these essential features from the channels composing the images is required; \textbf{(iii)}~traditional downscale and upscale operations rely on translational invariance and interpolation techniques, which are not able to learn a custom process for different tasks; \textbf{(iv)}~the module captures salient structure from the character fonts of the \gls*{lp}, highlighting both structure and textural features in the~image.

\begin{figure*}[!htb]
    \centering
    \captionsetup[subfigure]{captionskip=-0.25pt}
    
    \resizebox{0.975\linewidth}{!}{
    \subfloat[\phantom{-i}\label{subfig-1:dummy}]{%
        \includegraphics[height=32ex]{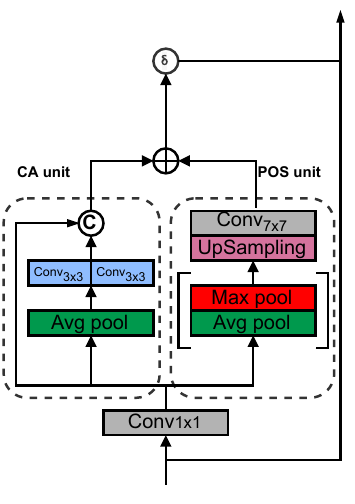}
    } \qquad
    \subfloat[\phantom{-i,}\label{subfig-2:dummy}]{%
        \includegraphics[height=32ex]{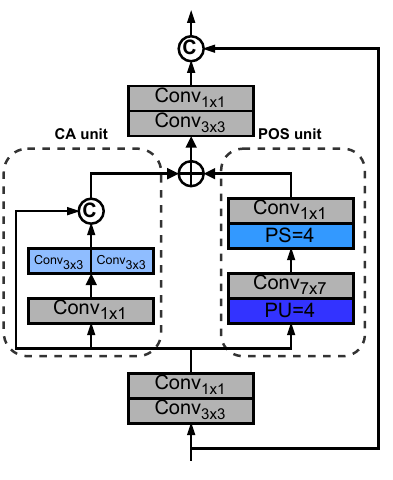}
    } \qquad
    \subfloat[\phantom{-i,}\label{subfig-3:dummy}]{%
        \includegraphics[height=32ex]{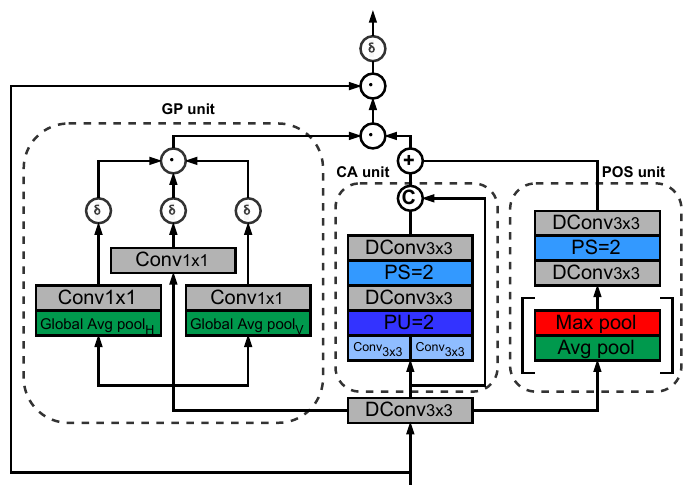}
    }
    }
    
    \vspace{-1mm}
    
    \caption{Comparative illustration of the (a)~Two-Fold Attention Module in MPRNet~\cite{mehri2021mprnet}, (b)~PixelShuffle Two-Fold Attention Module in Nascimento et al.~\cite{nascimento2022combining}, and (c)~PixelShuffle Three-Fold Attention Module~(ours). 
}
    \label{fig:pltfam}
\end{figure*}

The \gls*{ca} module is designed to identify and retain the inter-channel relationship features while eliminating less relevant ones.
This is accomplished by using two parallel convolutional layers, concatenating their outputs, and processing the combined output through a convolutional layer, a \gls*{pu} layer, a \gls*{ps} layer, and a \gls*{dconv} later.
This effectively summarizes the inter-channel relationship features for enhanced image~restoration.

The \gls*{pos} complements the \gls*{ca} module by identifying the location of significant features within the image. This is done by extracting first-order statistics through average and max pooling operations, concatenating the results, and processing them through \glspl*{dconv} and \gls*{ps} layers, restoring the original feature map dimension.
This highlights the positions of the relevant inter-channel relationship features, resulting in further improvement of image~restoration.

We incorporated a third branch named \gls*{gp} to the network to enhance its ability to extract critical characteristics such as structural, textural, and geometric features from the \gls{lp}.
This approach was motivated by the work of~\cite{zhang2023single}.
The \gls*{gp} utilizes global average pooling in both the vertical and horizontal directions of the input image.
The output from this layer is then subjected to a point-wise convolutional layer, followed by the sigmoid function to ensure the right channel dimensions.
The results from this layer are then aggregated through an element-wise multiplication to obtain the final output.

Finally, the outputs from the \gls*{ca}, \gls*{pos} and \gls*{gp} units are combined through an element-wise sum and multiplication to generate the final attention mask.
This mask summarizes all relevant information extracted by the \gls*{ca}, \gls*{pos} and \gls*{gp} units, and is used to enhance the input to the \gls*{pltfam} module through a \gls*{dconv} layer and a sigmoid function.
This process effectively emphasizes the key features of the image, including the inter-channel relationships, positional information, and structural information, resulting in improved image~restoration.

The original \glspl*{rcb} were enhanced by incorporating the proposed \gls{pltfam} instead of the \gls*{tfam} and including dilated convolution layers in the bottleneck path of the Adaptive Residual Blocks (ARB).
This modification retained the overall structure described in~\cite{mehri2021mprnet} while improving the network's capability to consider a broader context through an increased receptive field without adding extra parameters.
Also, the use of dilated convolutions helped to reproduce fine details in \gls*{lp} images by avoiding the ``smoothing'' effect that can occur with traditional~convolutions.

Returning our attention to \cref{fig:nnarch}, the reconstruction module was added as an output block for better aggregating fine details.
It consists of two \gls*{ps} with a scale factor value of $2$ for pixel reorganization, each followed by a \gls*{dconv} layer and by consecutive~\glspl*{rdb}.

\subsection{Perceptual Loss}

To further enhance the accuracy of \gls*{lpr}, we propose incorporating a perceptual loss function in our super-resolution approach. This loss function, shown in \cref{eq:loss}, is specifically designed to improve the accuracy of the system by considering the features that an \gls{ocr} model typically expects.

\begin{equation}
    PL=\frac{1}{n}\Big(\sum_{i=1}^{n}(H_i-S_i)^{2} + \sum_{i=1}^{n}|f_{OCR}(H_i)- f_{OCR}(S_i)|\Big)
    \label{eq:loss}
\end{equation}

In \cref{eq:loss}, $H_i$ and $S_i$ denote the high-resolution and super-resolved \gls*{lp} images, respectively, and $f_{OCR}(\cdot)$ represents the feature extraction process performed by the \gls{ocr}~model.

It is worth noting that the loss function allows the use of any \gls*{ocr} model for \gls*{lpr}.
This flexibility is particularly appealing since novel models can be readily incorporated as they become available.
In this work, we explored the multi-task model proposed by Gon\c{c}alves et al.~\cite{goncalves2018realtime}, as it is quite efficient and has achieved remarkable outcomes in prior research~\cite{goncalves2019multitask,nascimento2022combining}.

The \gls*{mse} is used to compute the difference between the expected and generated pixel values, with more significant errors being penalized more than minor errors.
This approach is beneficial in enhancing the overall quality of the image.
Also, the \gls*{mse} effectively preserves the structural information in the image, which is essential in the super-resolution task.
By contrast, the L1 loss ensures robustness to noise and outliers and helps to preserve sharp edges in the generated images by considering the expected features.
Combining \gls*{mse} and L1 loss allows for a more comprehensive evaluation of the generated images and helps achieve a balance between preserving structural information and minimizing~errors.
\section{Experiments}
\label{Experiments}

In this section, we detail the steps taken to validate the effectiveness of our proposed method for \gls*{lp} super-resolution.
We first describe our experimental setup and then proceed to provide a comprehensive analysis of the results~obtained.

\subsection{Setup}
\label{subsec:setup}

We made use of \gls*{lp} images obtained from the \rodosol~\cite{laroca2022cross} and \pku~\cite{yuan2017robust} datasets.
To the best of our knowledge, there is currently no public dataset that provides paired \gls*{lr} and \gls*{hr} images from real-world settings.
Hence, we opted for these two datasets since they provide a wide range of scenarios under which the images were~acquired.

\major{\rodosol is the largest public dataset acquired in Brazil.
It comprises $20{,}000$ images, with $10{,}000$ showing vehicles with Brazilian \glspl*{lp} and $10{,}000$ featuring vehicles with Mercosur \glspl*{lp}\footnote{Following~\cite{laroca2022first,nascimento2022combining,silva2022flexible}, we use the term ``Brazilian'' to refer to the layout used in Brazil prior to the adoption of the Mercosur~layout.}.}
Observe in \cref{fig:samples-rodosol} the diversity of this dataset regarding several factors such as \gls*{lp} colors, lighting conditions, and character fonts.
Here, we follow the standard protocol (defined in~\cite{laroca2022cross}) that involves using $40$\% of the images for training, $20$\% for validation, and $40$\% for~testing.

\begin{figure}[!htb]
    \centering
    
    \resizebox{0.825\linewidth}{!}{
    \includegraphics[width=0.31\linewidth, height=0.15\linewidth]{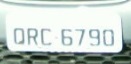} \hspace{-0.7mm}
    \includegraphics[width=0.31\linewidth, height=0.15\linewidth]{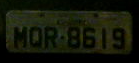} \hspace{-0.7mm}
    \includegraphics[width=0.31\linewidth, height=0.15\linewidth]{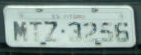} \hspace{-0.7mm}
    \includegraphics[width=0.31\linewidth, height=0.15\linewidth]{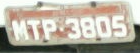}
    }
    
    \vspace{0.4mm}
    
    \resizebox{0.825\linewidth}{!}{
    \includegraphics[width=0.31\linewidth, height=0.15\linewidth]{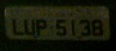} \hspace{-0.7mm}
    \includegraphics[width=0.31\linewidth, height=0.15\linewidth]{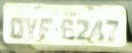} \hspace{-0.7mm}
    \includegraphics[width=0.31\linewidth, height=0.15\linewidth]{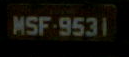} \hspace{-0.7mm}
    \includegraphics[width=0.31\linewidth, height=0.15\linewidth]{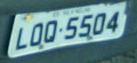}
    }
    
    \vspace{0.4mm}
    
    \resizebox{0.825\linewidth}{!}{
    \includegraphics[width=0.31\linewidth, height=0.15\linewidth]{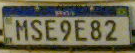} \hspace{-0.7mm}
    \includegraphics[width=0.31\linewidth, height=0.15\linewidth]{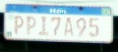} \hspace{-0.7mm}
    \includegraphics[width=0.31\linewidth, height=0.15\linewidth]{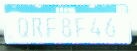} \hspace{-0.7mm}
    \includegraphics[width=0.31\linewidth, height=0.15\linewidth]{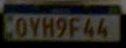}
    }
    
    \vspace{0.4mm}
    
    \resizebox{0.825\linewidth}{!}{
    \includegraphics[width=0.31\linewidth, height=0.15\linewidth]{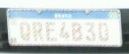} \hspace{-0.7mm}
    \includegraphics[width=0.31\linewidth, height=0.15\linewidth]{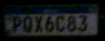} \hspace{-0.7mm}
    \includegraphics[width=0.31\linewidth, height=0.15\linewidth]{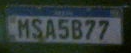} \hspace{-0.7mm}
    \includegraphics[width=0.31\linewidth, height=0.15\linewidth]{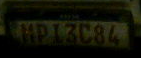}
    }
    
    \vspace{-2.75mm}
    
    \caption{Some \gls*{lp} images from the \rodosol dataset~\cite{laroca2022cross}. The first two rows show Brazilian \glspl*{lp}, while the last two rows show Mercosur \glspl*{lp}.
    For scope reasons, in this work, we conduct experiments on \glspl*{lp} that have all characters arranged in a single row (i.e., 10K images).
    }
    \label{fig:samples-rodosol}
\end{figure}

The \pku dataset comprises images categorized into five distinct groups, namely G1 through G5, each representing a specific scenario.
For instance, the images in G1 were captured on highways during the day and depict a single vehicle.
On the other hand, the images in G5 were taken at crosswalk intersections, either during the day or night, and have multiple vehicles.
All images were collected in mainland China.
\major{We perform experiments using the $2{,}253$ images in groups G1--G3, as they have labels regarding the \gls*{lp} text (these annotations were provided in~\cite{zhang2021robust_attentional}}).
Despite the diverse settings, the \gls*{lp} images have good quality and are perfectly legible (see some examples in \cref{fig:samples-PKU}).
Following \cite{zhang2021robust_attentional,laroca2022first}, we use $60$\% of the images for training/validation, while the remaining $40$\% are used for testing.
\major{Laroca et al.~\cite{laroca2023do} recently revealed that the \pku dataset (as well as several other datasets but not \rodosol) has multiple images of the same vehicle/\gls*{lp}.
They referred to such images as \emph{near-duplicates}.
Accordingly, to prevent bias in our experiments, we ensured that all images showing the same \gls*{lp} were grouped in the same~subset.}

\begin{figure}[!htb]
    \centering
    
    \resizebox{0.825\linewidth}{!}{
    \includegraphics[width=0.45\linewidth, height=0.15\linewidth]{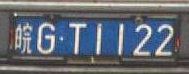} \hspace{-0.5mm}
    \includegraphics[width=0.45\linewidth, height=0.15\linewidth]{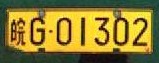} \hspace{-0.5mm}
    \includegraphics[width=0.45\linewidth, height=0.15\linewidth]{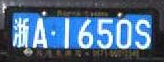}
    }
    
    \vspace{0.45mm}
    
    \resizebox{0.825\linewidth}{!}{
    \includegraphics[width=0.45\linewidth, height=0.15\linewidth]{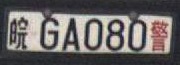} \hspace{-0.5mm}
    \includegraphics[width=0.45\linewidth, height=0.15\linewidth]{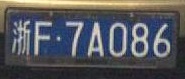} \hspace{-0.5mm}
    \includegraphics[width=0.45\linewidth, height=0.15\linewidth]{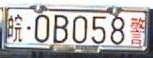}
    }
    
    \vspace{0.45mm}
    
    \resizebox{0.825\linewidth}{!}{
    \includegraphics[width=0.45\linewidth, height=0.15\linewidth]{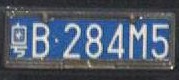} \hspace{-0.5mm}
    \includegraphics[width=0.45\linewidth, height=0.15\linewidth]{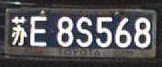} \hspace{-0.5mm}
    \includegraphics[width=0.45\linewidth, height=0.15\linewidth]{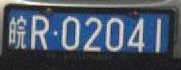}
    }
    
    \vspace{0.45mm}
    
    \resizebox{0.825\linewidth}{!}{
    \includegraphics[width=0.45\linewidth, height=0.15\linewidth]{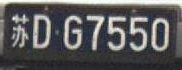} \hspace{-0.5mm}
    \includegraphics[width=0.45\linewidth, height=0.15\linewidth]{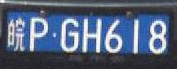} \hspace{-0.5mm}
    \includegraphics[width=0.45\linewidth, height=0.15\linewidth]{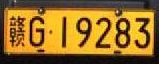}
    }
    
    \vspace{-2.75mm}
    
    \caption{Examples of \gls*{lp} images from the \pku dataset~\cite{yuan2017robust}. Although the \glspl*{lp} in this dataset have varying layouts, they all have seven characters.}
    \label{fig:samples-PKU}
\end{figure}

The \gls*{hr} images used in our experiments were generated as follows.
For each image from the chosen datasets, we first cropped the \gls*{lp} region using the annotations provided by the authors.
Afterward, we used the same annotations to rectify each \gls*{lp} image so that it becomes more horizontal, tightly bounded, and easier to recognize.
The rectified image is the \gls*{hr}~image.

Inspired by~\cite{zhang2018residual}, we generated \gls*{lr} versions of each \gls*{hr} image by simulating the effects of an optical system with lower resolution.
This was achieved by iteratively applying random Gaussian noise to each \gls*{hr} image until we reached the desired degradation level for a given \gls*{lr} image (i.e., \gls*{ssim}~$<0.1$).
To maintain the aspect ratio of the \gls*{lr} and \gls*{hr} images, \major{we padded them before resizing them to $20\times40$ pixels}, resulting in an output shape of $80\times160$ pixels for an upscale factor of~$4$.
\cref{fig:SSIM_examples} and \cref{fig:SSIM_examples_PKU} show examples of the \gls*{lp} images generated for the \rodosol and \pku datasets,~respectively.

\begin{figure}[!htb]
    \centering
    
    \resizebox{0.825\linewidth}{!}{
    \includegraphics[width=0.31\linewidth, height=0.15\linewidth]{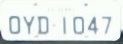} \hspace{-0.5mm}
    \includegraphics[width=0.31\linewidth, height=0.15\linewidth]{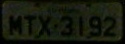} \hspace{-0.5mm}
    \includegraphics[width=0.31\linewidth, height=0.15\linewidth]{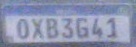} \hspace{-0.5mm}
    \includegraphics[width=0.31\linewidth, height=0.15\linewidth]{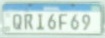} 
    }
    
    \vspace{0.6mm}
    
    \resizebox{0.825\linewidth}{!}{
    \includegraphics[width=0.31\linewidth, height=0.15\linewidth]{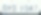} \hspace{-0.5mm}
    \includegraphics[width=0.31\linewidth, height=0.15\linewidth]{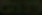} \hspace{-0.5mm}
    \includegraphics[width=0.31\linewidth, height=0.15\linewidth]{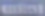} \hspace{-0.5mm}
    \includegraphics[width=0.31\linewidth, height=0.15\linewidth]{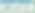}
    }    
    
    \vspace{-2.5mm}
    
    \caption{Some \gls*{hr}-\gls*{lr} image pairs created from the \rodosol~dataset.}
    \label{fig:SSIM_examples}
\end{figure}

\begin{figure}[!htb]
    \centering
    
    \resizebox{0.825\linewidth}{!}{
    \includegraphics[width=0.45\linewidth, height=0.15\linewidth]{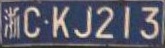} \hspace{-0.5mm}
    \includegraphics[width=0.45\linewidth, height=0.15\linewidth]{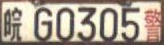} \hspace{-0.5mm}
    \includegraphics[width=0.45\linewidth, height=0.15\linewidth]{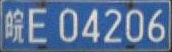} \hspace{-0.5mm}
    \includegraphics[width=0.45\linewidth, height=0.15\linewidth]{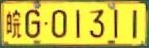} 
    }
    
    \vspace{0.6mm}
    
    \resizebox{0.825\linewidth}{!}{
    \includegraphics[width=0.45\linewidth, height=0.15\linewidth]{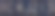} \hspace{-0.5mm}
    \includegraphics[width=0.45\linewidth, height=0.15\linewidth]{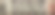} \hspace{-0.5mm}
    \includegraphics[width=0.45\linewidth, height=0.15\linewidth]{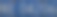} \hspace{-0.5mm}
    \includegraphics[width=0.45\linewidth, height=0.15\linewidth]{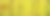}
    }    
    
    \vspace{-2.5mm}
    
    \caption{Examples of \gls*{hr}-\gls*{lr} image pairs created from the \pku dataset.}
    \label{fig:SSIM_examples_PKU}
\end{figure}

Our experiments were conducted using the PyTorch framework on a high-performance computer that is equipped with an AMD Ryzen $9$ $5950$X~CPU, $128$~GB of RAM, and an NVIDIA Quadro RTX $8000$ GPU ($48$~GB).

We used the Adam optimizer with a learning rate of $10$\textsuperscript{-$4$}, which decreases by a factor of $0.3$ (up to $10$\textsuperscript{-$7$}) when no improvement in the loss function is observed.
The training process stops after $20$ epochs without a decrease in the loss~function.

\subsection{Experimental Results}
\label{subsec:results}

In the \gls*{lpr} literature, models are usually evaluated in terms of the number of correctly recognized \glspl*{lp} divided by the number of \glspl*{lp} in the test set~\cite{wang2022rethinking,silva2022flexible,laroca2023do}.
A correctly recognized \gls*{lp} means that all characters on the \gls*{lp} were correctly recognized.
Considering our focus on low-resolution \glspl*{lp}, which are very common in forensic applications, we also report the recognition results considering partial matches (when at least $5$ or $6$ of the $7$ characters are correctly recognized)
as they may be useful in narrowing down the list of candidate \glspl*{lp} by incorporating additional information such as the vehicle's make and~model.

The results of the \gls{lpr} experiment are shown in Table~\ref{tab:ExperimentalResults}. The table demonstrates the recognition accuracy of \gls{hr} and \gls{lr} license plate images degraded by bicubic downsampling and recursive Gaussian noise. The difficulty of the task can be seen from the \gls{ssim} score, which ranges from $0$ to $0.10$, as illustrated in \cref{fig:SSIM_examples}, where the \gls*{lp} characters are barely~distinguishable.

\begin{table}[!htb]
\centering
\setlength{\tabcolsep}{2pt}
\renewcommand{\arraystretch}{1.05}
\caption{Recognition rates (\%) achieved in our experiments. ``All'' refers to \glspl*{lp} where all characters were recognized correctly; $\ge$~6 and $\ge$~5 refer to \glspl*{lp} where at least 6 or 5 characters were recognized correctly,~respectively.}

\vspace{-1.75mm}

\resizebox{0.95\linewidth}{!}{
\begin{tabular}{lcccccccccc}
\toprule
& & \multicolumn{3}{c}{\rodosol} & & & & \multicolumn{3}{c}{\pku}   \\
& & All & $\geq6$ & $\geq5$ & & & & All & $\geq6$ & $\geq5$   \\
\midrule                
\multicolumn{11}{l}{\gls*{ocr}~\cite{goncalves2018realtime} --- no super-resolution}\\
\midrule
\indentresults \gls*{hr} & & 96.6 & 98.6 & 99.0 & & & & 99.4 & 99.9 & 99.9  \\
\indentresults \gls*{lr}      & & \phantom{0}0.8 & \phantom{0}4.6 & 12.7 & & & & \phantom{0}0.0 &   \phantom{0}0.0 & \phantom{0}0.0 \\
\midrule
\midrule
\multicolumn{11}{l}{\gls*{ocr}~\cite{goncalves2018realtime} --- with super-resolution}\\
\midrule
\indentresults \textbf{Proposed}    & & \textbf{39.0} & \textbf{59.9} & \textbf{74.2} & & & & \textbf{72.0} & \textbf{90.3} & \textbf{97.3} \\
\indentresults Nascimento et al.~\cite{nascimento2022combining}   & & 10.5 & 25.4 & 42.2 & & & & 35.5 & 65.3 & 82.5 \\
\indentresults Mehri et al.~\cite{mehri2021mprnet}   & & 1.45 & 7.0 & 17.4 & & & & 22.5 & 49.2 & 70.6 \\
\midrule
\multicolumn{11}{l}{Average PSNR~(dB) and SSIM}\\ \midrule
  & & & PSNR & SSIM & & & &  & PSNR & SSIM  \\
\indentresults \textbf{Proposed} & & & 21.2 & \textbf{0.59}  & & & & & \textbf{18.3} & \textbf{0.61} \\
\indentresults Nascimento et al.~\cite{nascimento2022combining} & & & \textbf{21.3} & 0.52 &  & & & & 18.1 & 0.54 \\
\indentresults Mehri et al.~\cite{mehri2021mprnet} & & & 16.8 & 0.38 & & & & & 16.4 & 0.41 \\

\bottomrule
\end{tabular} \,
}

\label{tab:ExperimentalResults}
\end{table}

The proposed super-resolution network achieved superior performance compared to the two baseline models~\cite{mehri2021mprnet,nascimento2022combining}, as presented in the second section of Table~\ref{tab:ExperimentalResults}.
The multi-task \gls*{ocr} model~\cite{goncalves2018realtime} demonstrated remarkable improvement when applied to images reconstructed by our super-resolution approach in both datasets, particularly in the \pku dataset, with a $14.8\%$ higher recognition rate compared to the method proposed in our preliminary method~\cite{nascimento2022combining} and a $26.7\%$ higher accuracy compared to \gls*{mprnet}~\cite{mehri2021mprnet} for \glspl*{lp} with more than five correct characters.

For completeness, we detail in \cref{tab:ExperimentalResults} the \gls*{psnr} and \gls*{ssim} obtained by each approach.
Similar to what was observed in~\cite{zhang2018unreasonable,hamdi2021new,lin2021license}, the \gls*{psnr} metric seems inappropriate for this particular application, as our approach and the one proposed in \cite{nascimento2022combining} reached comparable values, despite ours leading to significantly better results achieved by the \gls*{ocr} model.
The \gls*{ssim} metric, on the other hand, seems to better represent the quality of reconstruction of \gls*{lp} images, as the proposed method achieved considerably better \gls*{ssim} values in both~datasets.

The variation in accuracy between the two datasets can be attributed to the diversity present in the \rodosol dataset, which includes a range of layouts, lighting conditions, and character fonts, while the \pku dataset largely comprises \glspl*{lp} with a uniform layout, with less variation in the environmental conditions under which the images were~collected.

The \gls*{ocr} network showed these improved results because of the effective extraction of textural and structural information by the proposed \gls*{gp} unit, along with the \gls*{ca} and \gls*{pos} units. These units were designed with pyramid and PixelShuffle layers to optimize channel scaling and reorganization within the~image.

Finally, the results of the \gls*{lpr} experiments are further substantiated by a visual contrast of the super-resolution images produced by our technique and the baseline methods~\cite{mehri2021mprnet,nascimento2022combining}.
\cref{fig:Qresults} and~\cref{fig:QresultsPKU} show four \gls*{lr} images alongside their corresponding super-resolution counterparts, in addition to the original \gls*{hr} image as a reference.
It is evident that the proposed approach outperforms both its preliminary version~\cite{nascimento2022combining} and \gls*{mprnet}~\cite{mehri2021mprnet} in terms of perceptual~quality.

\begin{figure}[!htb]
    \captionsetup[subfigure]{labelformat=empty,position=top,captionskip=0.75pt,justification=centering} 

    \vspace{0.7mm}
    
    \centering
    \resizebox{0.9\linewidth}{!}{
    \subfloat[LR (Input)]{
    \includegraphics[width=0.29\linewidth, height=0.145\linewidth]{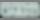}
    } \hspace{-2.00mm}
    \subfloat[Mehri et al.~\cite{mehri2021mprnet}]{
    \includegraphics[width=0.29\linewidth, height=0.145\linewidth]{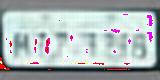}
    } \hspace{-2.00mm}
    \subfloat[Nascimento et al.~\cite{nascimento2022combining}]{
    \includegraphics[width=0.29\linewidth, height=0.145\linewidth]{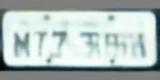}
    } \hspace{-2.00mm}
    \subfloat[Proposed]{
    \includegraphics[width=0.29\linewidth, height=0.145\linewidth]{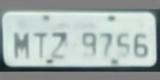}
    } \hspace{-2.00mm}
    \subfloat[HR (GT)]{
    \includegraphics[width=0.29\linewidth, height=0.145\linewidth]{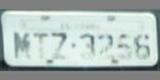} 
    } \,
    }
    
    \vspace{-2.15mm}
    
    \resizebox{0.9\linewidth}{!}{
    \subfloat[]{
    \includegraphics[width=0.29\linewidth, height=0.145\linewidth]{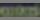}
    } \hspace{-2.00mm}
    \subfloat[]{
    \includegraphics[width=0.29\linewidth, height=0.145\linewidth]{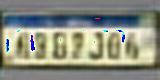} 
    } \hspace{-2.00mm}
    \subfloat[]{
    \includegraphics[width=0.29\linewidth, height=0.145\linewidth]{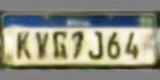} 
    } \hspace{-2.00mm}
    \subfloat[]{
    \includegraphics[width=0.29\linewidth, height=0.145\linewidth]{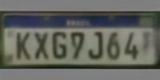} 
    } \hspace{-2.00mm}
    \subfloat[]{
    \includegraphics[width=0.29\linewidth, height=0.145\linewidth]{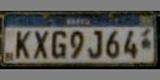} 
    } \,
    }
    
    \vspace{-2.15mm}

    \resizebox{0.9\linewidth}{!}{
    \subfloat[]{
    \includegraphics[width=0.29\linewidth, height=0.145\linewidth]{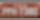}
    } \hspace{-2.00mm}
    \subfloat[]{
    \includegraphics[width=0.29\linewidth, height=0.145\linewidth]{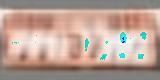}
    } \hspace{-2.00mm}
    \subfloat[]{
    \includegraphics[width=0.29\linewidth, height=0.145\linewidth]{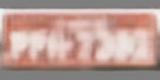}
    } \hspace{-2.00mm}
    \subfloat[]{
    \includegraphics[width=0.29\linewidth, height=0.145\linewidth]{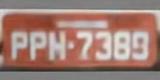}
    } \hspace{-2.00mm}
    \subfloat[]{
    \includegraphics[width=0.29\linewidth, height=0.145\linewidth]{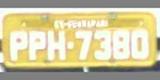} 
    } \,
    } 
    
    \vspace{-2.15mm}

    \resizebox{0.9\linewidth}{!}{
    \subfloat[]{
    \includegraphics[width=0.29\linewidth, height=0.145\linewidth]{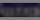}
    } \hspace{-2.00mm}
    \subfloat[]{
    \includegraphics[width=0.29\linewidth, height=0.145\linewidth]{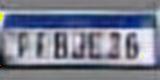}
    } \hspace{-2.00mm}
    \subfloat[]{
    \includegraphics[width=0.29\linewidth, height=0.145\linewidth]{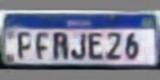}
    } \hspace{-2.00mm}
    \subfloat[]{
    \includegraphics[width=0.29\linewidth, height=0.145\linewidth]{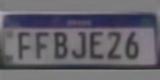}
    } \hspace{-2.00mm}
    \subfloat[]{
    \includegraphics[width=0.29\linewidth, height=0.145\linewidth]{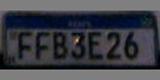} 
    } \,
    } 

    \vspace{-0.25mm}
    \caption{Typical examples of the images generated by the proposed approach and baselines in the \rodosol dataset~\cite{laroca2022cross}. GT = ground truth.}
    \label{fig:Qresults}
\end{figure}

\begin{figure}[!htb]
    \captionsetup[subfigure]{labelformat=empty,position=top,captionskip=0.75pt,justification=centering}
    
    \centering
    \resizebox{0.95\linewidth}{!}{
    \subfloat[LR (Input)]{
    \includegraphics[width=0.325\linewidth, height=0.10\linewidth]{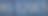}
    } \hspace{-2.00mm}
    \subfloat[Mehri et al.~\cite{mehri2021mprnet}]{
    \includegraphics[width=0.325\linewidth, height=0.10\linewidth]{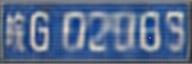}
    } \hspace{-2.00mm}
    \subfloat[Nascimento et al.~\cite{nascimento2022combining}]{
    \includegraphics[width=0.325\linewidth, height=0.10\linewidth]{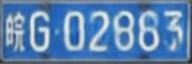}
    } \hspace{-2.00mm}
    \subfloat[Proposed]{
    \includegraphics[width=0.325\linewidth, height=0.10\linewidth]{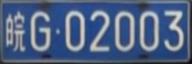}
    } \hspace{-2.00mm}
    \subfloat[HR (GT)]{
    \includegraphics[width=0.325\linewidth, height=0.10\linewidth]{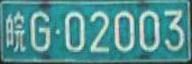} 
    } \,
    }

    \vspace{-2.00mm}

    \resizebox{0.95\linewidth}{!}{
    \subfloat[]{
    \includegraphics[width=0.325\linewidth, height=0.10\linewidth]{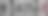}
    } \hspace{-2.00mm}
    \subfloat[]{
    \includegraphics[width=0.325\linewidth, height=0.10\linewidth]{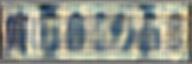}
    } \hspace{-2.00mm}
    \subfloat[]{
    \includegraphics[width=0.325\linewidth, height=0.10\linewidth]{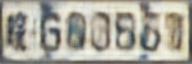}
    } \hspace{-2.00mm}
    \subfloat[]{
    \includegraphics[width=0.325\linewidth, height=0.10\linewidth]{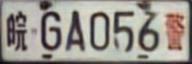}
    } \hspace{-2.00mm}
    \subfloat[]{
    \includegraphics[width=0.325\linewidth, height=0.10\linewidth]{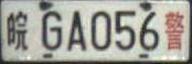} 
    } \,
    } 
    
    \vspace{-2.00mm}
     
    \resizebox{0.95\linewidth}{!}{
    \subfloat[]{
    \includegraphics[width=0.325\linewidth, height=0.10\linewidth]{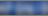}
    } \hspace{-2.00mm}
    \subfloat[]{
    \includegraphics[width=0.325\linewidth, height=0.10\linewidth]{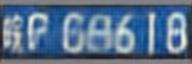}
    } \hspace{-2.00mm}
    \subfloat[]{
    \includegraphics[width=0.325\linewidth, height=0.10\linewidth]{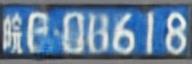}
    } \hspace{-2.00mm}
    \subfloat[]{
    \includegraphics[width=0.325\linewidth, height=0.10\linewidth]{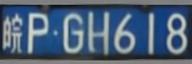}
    } \hspace{-2.00mm}
    \subfloat[]{
    \includegraphics[width=0.325\linewidth, height=0.10\linewidth]{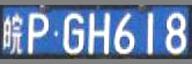} 
    } \,
    } 
    
    \vspace{-2.00mm}

    \resizebox{0.95\linewidth}{!}{
    \subfloat[]{
    \includegraphics[width=0.325\linewidth, height=0.10\linewidth]{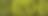}
    } \hspace{-2.00mm}
    \subfloat[]{
    \includegraphics[width=0.325\linewidth, height=0.10\linewidth]{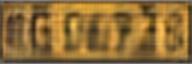} 
    } \hspace{-2.00mm}
    \subfloat[]{
    \includegraphics[width=0.325\linewidth, height=0.10\linewidth]{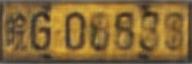} 
    } \hspace{-2.00mm}
    \subfloat[]{
    \includegraphics[width=0.325\linewidth, height=0.10\linewidth]{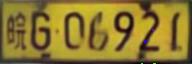} 
    } \hspace{-2.00mm}
    \subfloat[]{
    \includegraphics[width=0.325\linewidth, height=0.10\linewidth]{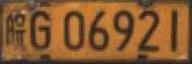} 
    } \,
    }
    
    \vspace{-2.00mm}

    \resizebox{0.95\linewidth}{!}{
    \subfloat[]{
    \includegraphics[width=0.325\linewidth, height=0.10\linewidth]{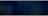}
    } \hspace{-2.00mm}
    \subfloat[]{
    \includegraphics[width=0.325\linewidth, height=0.10\linewidth]{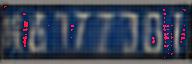}
    } \hspace{-2.00mm}
    \subfloat[]{
    \includegraphics[width=0.325\linewidth, height=0.10\linewidth]{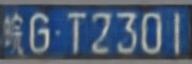}
    } \hspace{-2.00mm}
    \subfloat[]{
    \includegraphics[width=0.325\linewidth, height=0.10\linewidth]{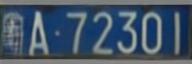}
    } \hspace{-2.00mm}
    \subfloat[]{
    \includegraphics[width=0.325\linewidth, height=0.10\linewidth]{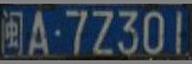} 
    } \,
    } 
  
    \vspace{-0.25mm}
    
    \caption{Representative samples of the images generated by the proposed approach and baselines in the \pku dataset~\cite{yuan2017robust}. GT = ground truth.}
    \label{fig:QresultsPKU}
\end{figure}

In general, the images produced by \gls*{mprnet}~\cite{mehri2021mprnet} exhibit a common issue of blurriness, where the character edges blend into the \gls*{lp} background, resulting in visible artifacts.
This blurriness can also cause the edges of multiple characters to blend together, leading to further visual distortions.
The architecture proposed in our previous work~\cite{nascimento2022combining} manages to reconstruct the characters but distorts them with strong undulations, making them appear as part of the \gls*{lp} background in some cases (see the first row of \cref{fig:Qresults}).
Conversely, the proposed model generates clear character edges and consistently reconstructs the original font, without any missing characters or incomplete~lines.

It is notable that when our model is uncertain about which character to reconstruct, it tends to hallucinate with characters that are most congruent with the \gls*{lr} input, as evident in the last row of~\cref{fig:Qresults} and~\cref{fig:QresultsPKU}, where the character ``3'' is reconstructed as ``J'', and the character ``Z'' is reconstructed as ``2'', respectively.
\major{This issue can be mitigated by incorporating a lexicon or vocabulary into the network's learning process to track the character type (letter, digit, or either) that can occupy each position on \glspl*{lp} of a specific~layout.}

Furthermore, the network tends to generate nearly identical background colors for different images.
This behavior can be observed in the third row of \cref{fig:Qresults} and the first row of \cref{fig:QresultsPKU}.
\major{However, it is noteworthy that, based on our analysis, this does not considerably impact the recognition results~achieved.}

\major{\subsubsection{Ablation Study}}

\major{
As the proposed approach integrates multiple concepts into a single architecture, we conducted an ablation study to validate the contribution of each incorporated unit to the results obtained.
The study involved removing the autoencoder, \gls*{tfam}, \gls*{ps} and \gls*{pu} layers and training the network without the perceptual loss (one modification at a time).
}

\major{Four baselines were established for the experiments.
The first baseline replaced the autoencoder with a \gls*{dconv} layer with a $5\times5$ kernel for shallow feature extraction~\cite{mehri2021mprnet}.
The second baseline removed the TFAM module and adjusted the output of the previous layer to match the input shape of the following layers.
The third baseline replaced the PS and PU layers with transposed and strided convolution layers, respectively, as they are analogous~\cite{shi2016realtime}. 
Finally, in the fourth baseline, the perceptual loss was replaced by \gls*{mse}, which is commonly used in super-resolution research~\cite{wang2021deep, liu2023blind}.
\cref{tab:ab-table} presents the~results.
}

\begin{table}[!htb]
\centering
\caption{\major{Recognition rates (\%) achieved in the ablation study. ``All'' refers to \glspl*{lp} where all characters were recognized correctly; $\ge$~6 and $\ge$~5 refer to \glspl*{lp} where at least 6 or 5 characters were recognized correctly,~respectively.}}
\label{tab:ab-table}

\vspace{-1.75mm}

\resizebox{0.975\linewidth}{!}{
\major{
\begin{tabular}{lcccccc}
\toprule
\multirow{2}{*}{Approach} & \multicolumn{3}{c}{\rodosol} & \multicolumn{3}{c}{\pku} \\
 & All & $\geq6$ & $\geq5$ & All & $\geq6$ & $\geq5$ \\ \midrule
Proposed (w/o autoencoder) & $32.7$ & $55.0$ & $70.1$ & $\textbf{73.8}$ & $90.2$ & $96.6$ \\
Proposed (w/o \gls*{tfam}) & $33.3$ & $55.0$ & $69.6$ & $73.1$ & $90.1$ & $96.6$ \\
Proposed (w/o \gls*{ps} and \gls*{pu} layers) & $34.3$ & $54.8$ & $68.5$ & $70.4$ & $89.9$ & $96.7$ \\
Proposed (w/o perceptual loss) & $35.6$ & $57.3$ & $71.9$ & $72.4$ & $\textbf{91.4}$ & $97.1$ \\ \midrule
Proposed & $\textbf{39.0}$ & $\textbf{59.9}$ & $\textbf{74.2}$ & $72.0$ & $90.3$ & $\textbf{97.3}$ \\ \bottomrule
\end{tabular} \,
}
}
\end{table}

\major{The results of the experiments on the \rodosol dataset demonstrate that each of the units included in the proposed system significantly contributes to its overall performance.
The complete system attained a recognition rate of~$39.0$\%, while the best version without one of the components reached a recognition rate of~$35.6$\%.
The worst-case scenario was when the autoencoder unit was removed, resulting in a recognition rate of $32.7$\% for all characters recognized.
This is because the autoencoder module plays a vital role in facilitating the extraction of shallow features.
Specifically, the autoencoder generates a mask by squeezing and expanding the input image, highlighting the most critical areas for reconstruction by the rest of the network.
Without this mask, the network struggles to identify the relevant features, resulting in poor~performance.}

\major{
In contrast, the recognition rates in the \pku dataset were only enhanced with the incorporation of \gls*{ps} and \gls*{pu} layers.
We conjecture that the other units are not required for this dataset due to its images being considerably less complex than those in the \rodosol (as evidenced by the images in \cref{fig:samples-rodosol} and \cref{fig:samples-PKU}).
This could explain why several authors opted to conduct ablation studies solely on the largest and most diverse dataset among those used in their experiments~\cite{zhang2021robust_attentional,qin2022towards,wang2022rethinking}.}
\section{Conclusions}
\label{Conclusion}

This article proposes a new super-resolution approach to improve the recognition of low-resolution \glspl*{lp}.
Our method builds upon the existing \gls{mprnet}~\cite{mehri2021mprnet} and the architecture proposed in our previous work~\cite{nascimento2022combining} by incorporating subpixel-convolution layers (\gls{ps} and \gls{pu}) in combination with a \gls{pltfam}.
We also introduce a novel perceptual loss that combines features extracted from an \gls*{ocr} model with L1 loss to reconstruct characters with the most relevant characteristics, while also incorporating \gls{mse} to enhance overall image~quality.

Our approach capitalizes on both structural and textural features by using the \gls*{ps} and \gls*{pu} layers for custom scale operations, rather than relying on conventional translational invariance and interpolation techniques.
An autoencoder with \gls*{ps} and \gls*{pu} layers was integrated to extract shallow features and generate an attention mask that is added to the original input.
The output of the autoencoder is processed by a \gls*{rdb} to identify regions of interest for reconstruction, optimizing computational resources and producing super-resolution images that emphasize relevant~information.

We conducted experiments on two publicly available datasets from Brazil and mainland China, which contain a diverse range of \gls*{lp} images.
The results showed better recognition rates being achieved in the images reconstructed by the proposed method than in those reconstructed by the  baselines.
More specifically, for the \rodosol dataset, our method led to a recognition rate of $39.0$\% being achieved by the \gls*{ocr} model, while the methods proposed in \cite{nascimento2022combining} and \cite{mehri2021mprnet} led to recognition rates of $31.3$\% and $4.0$\%, respectively.
Similarly, for the \pku dataset, our approach outperformed both baselines, with the \gls*{ocr} model reaching a recognition rate of $72.0\%$, compared to $35.5\%$ and $22.5\%$ for \cite{nascimento2022combining} and \cite{mehri2021mprnet}, respectively.
\major{We have made available all datasets used in our experiments (i.e., the LR--HR image pairs), as well as the source code, in order to encourage further research and development in the field of \gls*{lpr}~super-resolution.}

\major{In the future, our plans include integrating a lexicon or vocabulary into the network's learning process to track the character type that can occupy each position on \glspl*{lp} of a specific layout.
Additionally,} we intend to create a large-scale dataset for \gls*{lp} super-resolution, consisting of thousands of \gls*{lr} and \gls*{hr} image pairs.
We aim to collect videos in which the \gls*{lp} is legible in one frame but not in another, enabling us to assess existing methods in real-world scenarios and develop novel~methods. 

\section*{Acknowledgments}

This work was supported in part by the Coordination for the Improvement of Higher Education Personnel~(CAPES) (\textit{Programa de Coopera\c{c}\~{a}o Acad\^{e}mica em Seguran\c{c}a P\'{u}blica e Ci\^{e}ncias Forenses \#~88881.516265/2020-01}), in part by the National Council for Scientific and Technological Development~(CNPq) (\#~309953/2019-7 and \#~308879/2020-1), and also in part by the Minas Gerais Research Foundation (FAPEMIG) (Grant~PPM-00540-17).
We gratefully acknowledge the support of NVIDIA Corporation with the donation of the Quadro RTX $8000$ GPU used for this research.

\balance
\bibliographystyle{IEEEtran}
\bibliography{bibtex}
\fi
\end{document}